\documentclass{sig-alternate}

\usepackage{enumitem}% http://ctan.org/pkg/enumitem
\usepackage{pstricks}
\usepackage{graphicx}
\usepackage{hyperref}

\begin{document}
%
% --- Author Metadata here ---
\conferenceinfo{KDD}{2014 New York, New York USA}
%\CopyrightYear{2007} % Allows default copyright year (20XX) to be over-ridden - IF NEED BE.
%\crdata{0-12345-67-8/90/01}  % Allows default copyright data (0-89791-88-6/97/05) to be over-ridden - IF NEED BE.
% --- End of Author Metadata ---

\title{Machine Learning at Scale}

%
% You need the command \numberofauthors to handle the 'placement
% and alignment' of the authors beneath the title.
%
% For aesthetic reasons, we recommend 'three authors at a time'
% i.e. three 'name/affiliation blocks' be placed beneath the title.
%
% NOTE: You are NOT restricted in how many 'rows' of
% "name/affiliations" may appear. We just ask that you restrict
% the number of 'columns' to three.
%
% Because of the available 'opening page real-estate'
% we ask you to refrain from putting more than six authors
% (two rows with three columns) beneath the article title.
% More than six makes the first-page appear very cluttered indeed.
%
% Use the \alignauthor commands to handle the names
% and affiliations for an 'aesthetic maximum' of six authors.
% Add names, affiliations, addresses for
% the seventh etc. author(s) as the argument for the
% \additionalauthors command.
% These 'additional authors' will be output/set for you
% without further effort on your part as the last section in
% the body of your article BEFORE References or any Appendices.

\numberofauthors{2} %  in this sample file, there are a *total*
% of EIGHT authors. SIX appear on the 'first-page' (for formatting
% reasons) and the remaining two appear in the \additionalauthors section.
%
\author{
% You can go ahead and credit any number of authors here,
% e.g. one 'row of three' or two rows (consisting of one row of three
% and a second row of one, two or three).
%
% The command \alignauthor (no curly braces needed) should
% precede each author name, affiliation/snail-mail address and
% e-mail address. Additionally, tag each line of
% affiliation/address with \affaddr, and tag the
% e-mail address with \email.
%
% 1st. author
\alignauthor Sergei Izrailev\\
       \affaddr{Collective, Inc.}\\
       \affaddr{99 Park Ave}\\
       \affaddr{5th Floor}\\
       \affaddr{New York, NY 10016}\\
       \email{sizrailev@collective.com}
% 2nd. author
\alignauthor Jeremy M. Stanley\\
       \affaddr{Collective, Inc.}\\
       \affaddr{99 Park Ave}\\
       \affaddr{5th Floor}\\
       \affaddr{New York, NY 10016}\\
       \email{jstanley@collective.com}
}

% There's nothing stopping you putting the seventh, eighth, etc.
% author on the opening page (as the 'third row') but we ask,
% for aesthetic reasons that you place these 'additional authors'
% in the \additional authors block, viz.

%\additionalauthors{Additional authors: John Smith (The Th{\o}rv{\"a}ld Group,
%email: {\texttt{jsmith@affiliation.org}}) and Julius P.~Kumquat
%(The Kumquat Consortium, email: {\texttt{jpkumquat@consortium.net}}).}
%\date{30 July 1999}

% Just remember to make sure that the TOTAL number of authors
% is the number that will appear on the first page PLUS the
% number that will appear in the \additionalauthors section.

\maketitle
\begin{abstract}

It takes skill to build a meaningful predictive model even with the abundance of
implementations of modern machine learning algorithms and readily available
computing resources. Building a model becomes challenging if hundreds of
terabytes of data need to be processed to produce the training data set. In a
digital advertising technology setting, we are faced with the need to build
thousands of such models that predict user behavior and power advertising
campaigns in a 24/7 chaotic real-time production environment. As data
scientists, we also have to convince other internal departments critical to
implementation success, our management, and our customers that our machine
learning system works. In this paper, we present the details of the design and
implementation of an automated, robust machine learning platform that impacts
billions of advertising impressions monthly. This platform enables us to
continuously optimize thousands of campaigns over hundreds of millions of users,
on multiple continents, against varying performance objectives.

\end{abstract}

% A category with the (minimum) three required fields
%A category including the fourth, optional field follows...
%\category{D.2.8}{Software Engineering}{Metrics}[complexity measures,performance measures]



% introduction
\section{Introduction}

Demand for digital advertising from consumer brands has been growing rapidly in
the past few years because of a recent shift to more and more content becoming
accessible to people on digital devices. Digital advertising spans display
(banner), video and rich media ads that are shown to users while they are
browsing the Internet, as well as ads shown in mobile apps and on addressable TV
sets. In contrast with traditional TV advertising, which is generally reaching
households within the intended demographics and geographic location, digital
advertising offers the advantage of targeting specific users and measuring their
responses. This is frequently implemented by creating pools of browser cookies
that anonymously represent people within the target audience for an advertising
campaign. When a web page with space allocated for an ad loads in the web
browser, the browser generally sends a request for an ad along with the user's
browser cookie to an ad server. Part of the decision by the ad server of which
ad to send back to the browser depends on whether the cookie is in a targeted
pool.

A common way of generating pools of users is collecting cookies of people who
expressed interest in a given product or service by visiting the advertiser's
web site (known existing or potential customers). However, advertisers are also
interested in finding other people for whom their ad may be relevant (unknown
potential customers). The latter is frequently addressed by building a model
predicting which cookies belong to people who are in the target audience for a
given campaign \cite{m6d:bidopt,turn:convrate,turn:perfgoal,criteo:scalable}.

The target audience for a given advertising campaign can be defined in different
ways, largely depending on the campaign goals. In general, these goals are of
two types: reach and performance. Campaigns with reach goals are targeting a
certain audience segment, e.g., people likely to watch a given TV show, or
likely to be interested in sports. Campaigns with performance goals, on the
other hand, are targeting users who are likely to complete a certain action in
the future, e.g., click on the ad or purchase a product online after viewing the
ad.

At Collective we are faced with the challenges of building, testing, maintaining
and keeping track of thousands of such models that enable daily delivery of tens
of millions of ad impressions for hundreds of advertising campaigns. The data
involved occupies hundreds of terabytes of storage, it is time-dependent and
comes from multiple and evolving data sources in different formats. The
campaigns run continuously (24/7) on our platform and are routinely added,
removed or modified. Finally, the resulting user pools must be kept up-to-date
in the real-time systems that deliver ads. This paper describes the system
currently deployed in production at Collective that enables building and
maintaining thousands of predictive models at a time, and making hundreds of
billions of predictions on a daily basis.

\section{Model Setup}

In this section, we define the goals of our modeling platform, discuss the time
structure of the data used for model building, outline the features used to make
predictions and describe our choice of modeling methodology.

%-------------------------------------------------------------------------------

\subsection{Goals of Modeling}

The modeling needs of campaigns with performance and reach goals are very
similar, however, for simplicity we will only describe models for campaigns with
performance goals. When an ad is sent by the ad server to the user's browser (an
event referred to as an ``impression'') there is a chain of other events that
follows. First of all, the user may or may not see the ad, depending on whether
the ad is above or below the fold of the web page and is ``viewable''. A
standard for measuring ad viewability has emerged recently to mean ``50\% or
more of the ad was visible on the screen for 1 second or longer''
\cite{IAB_3ms}. If the ad is viewable, the user may interact with the ad in some
way. For example, the user may bring the mouse cursor to the ad, or, in case of
video ads, view it to completion or skip it. Then, the user may click on the ad
and follow the link to the advertiser's web page. Finally, the user may
``convert'', which can mean buying a product, signing up for an online
newsletter, requesting information for a product, etc. Note that conversions may
happen regardless of whether or not the user interacted with or clicked on the
ad.

In the chain of events, from the ad being available in the browser, to view, to
interaction, to click, to conversion, the chances of the desired event happening
go down, while the value of the event to the advertiser goes up. Every campaign
with performance goals has one or more goals within this chain of events. As a
result, we frequently have to build more than one type of model per campaign. In
addition, when the target events are too rare to build a meaningful model, which
may happen at the very beginning of a campaign with no prior history, the system
automatically falls back to a previous event type in the chain that has a higher
frequency of occurrence. For example, a model for predicting clicks would fall
back to a model predicting user interactions with an ad. 

In order to create a unified modeling process across all event types, we take
advantage of the fact that in all cases the response variable in the model can
be treated as a binary variable that identifies whether or not a given event
happened within a predefined time interval $T$ after the ad was served. Thus,
for each model we are predicting the probability of a given user $U$ to complete
a specific action $A$ within time $T$ (``look-forward window") after being
served an ad. This formulation can be readily extended to the response variable
representing counts of events, and the model predicting the expected number of
user actions.

%-------------------------------------------------------------------------------

\subsection{Time Structure of Data}

Because our models are predicting events occurring in the future, the response
variable has to be measured later in time relative to the state of the predictor
variables. Figure 1 shows a diagram of the time structure of the predictor and
response variables. The users are observed over a period of time, at the end of
which predictions are made and the users are assigned to the target user pools.
There is a short delay arising from the time taken to deliver the data to
real-time systems. Once the user pools are available for targeting on the ad
server, the response interval begins. The response interval accounts for all
occurrences of ad delivery and the events that are considered a success for the
given campaign within a look-forward window after an ad was delivered.

% Figure 1
\begin{figure}
\centering
\input{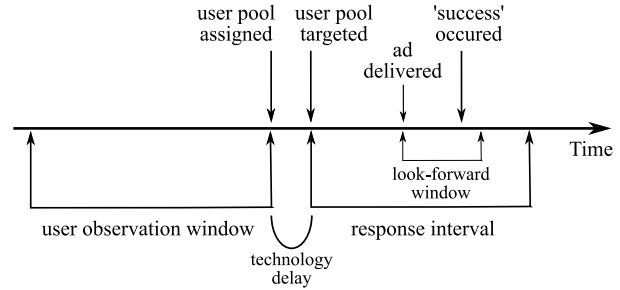}
\caption{Time relationship between predictor and response variables.}
\end{figure}

The same time structure of the data is reflected in the training data set, as
shown in Figure 2. We use the information that we knew about a user at
the end of the observation window as predictor variables, while the response
variable indicates whether or not the user has performed the desired action
given that we delivered an ad to this user. We construct a sequence of
non-overlapping response intervals of equal length, except that the latest
response interval may be incomplete in order for the model to take advantage of
the most recent available data. Note, however, that the look-forward interval
for capturing the response remains consistent to avoid biasing the outcome rate
in the most recent interval.

Typically, the user observation period is four weeks, the response interval is
one week, and we use a set of 8 such intervals to construct the training data
set. The number of response intervals is a tradeoff between the amount of
historical data we have to store, its relevance to the current user's behavior,
and the number of positive events available to the model. The data in the
response intervals is exponentially weighted so that more recent data
contributes more to the model than the older data. In addition, in cases where
the number of success events is very large, only events from the most recent
intervals are included in the training set.

% Figure 2
\begin{figure}
\centering
\scalebox{0.9}{\input{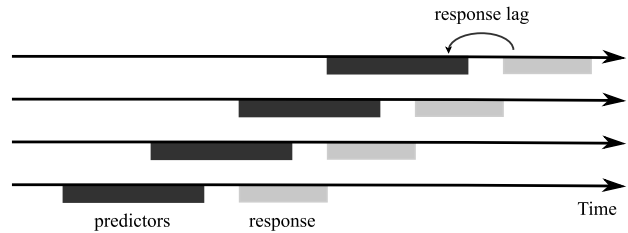}}
\caption{Time structure of the training data.}
\end{figure}

%-------------------------------------------------------------------------------

\begin{table*}
\centering
\caption{Strengths and limitations of \texttt{glmnet}}
\begin{tabular}{|l|l|l|} \hline
\multicolumn{1}{|c|}{Component} & 
\multicolumn{1}{|c|}{Strengths} & 
\multicolumn{1}{|c|}{Limitations}\\ \hline
Logistic regression model & 
\parbox{0.3\textwidth}{
\begin{itemize}[noitemsep,topsep=0pt,label=$\cdot$] 
  \item Produces probabilities$^*$ 
  \item Speed of predictions$^*$ 
  \item Interpretability of coefficients
\end{itemize}} & 
\parbox{0.3\textwidth}{
\begin{itemize}[noitemsep,topsep=0pt,label=$\cdot$]
  \item Nonlinearity must be specified$^*$ 
  \item Interactions must be specified$^*$ 
  \item Limited missing value support
\end{itemize}}\\ 
\hline 
Elastic net regularization & 
\parbox{0.3\textwidth}{
\begin{itemize}[noitemsep,topsep=0pt,label=$\cdot$] 
  \item Reduces risk of overfitting$^*$ 
  \item Parsimonious with $\alpha$ near 1
\end{itemize}} & 
\parbox{0.3\textwidth}{
\begin{itemize}[noitemsep,topsep=0pt,label=$\cdot$]
  \item Limits statistical inference
  \item Requires choosing $\alpha$
\end{itemize}}\\ 
\hline 
Coordinate gradient descent & 
\parbox{0.3\textwidth}{
\begin{itemize}[noitemsep,topsep=0pt,label=$\cdot$] 
  \item $\lambda$ search path for ``free''$^*$ 
  \item Sparse data support$^*$ 
\end{itemize}} & \\ 
\hline
R package implementation & 
\parbox{0.3\textwidth}{
\begin{itemize}[noitemsep,topsep=0pt,label=$\cdot$] 
  \item Cross-validation for $\lambda$ search
  \item Ease of use of R interface
  \item Speed of model building$^*$ 
\end{itemize}} & 
\parbox{0.3\textwidth}{
\begin{itemize}[noitemsep,topsep=0pt,label=$\cdot$]
  \item Not parallelized except for cross-validation
\end{itemize}}\\ 
\hline
\multicolumn{3}{l}{$^*$ These strengths (and limitations) are most significant,
and are explained in the text.}\\ 
\end{tabular}
\end{table*}

%-------------------------------------------------------------------------------

\subsection{Predictor Variables}
   
The data that is meaningful as features for predicting user behavior comes
primarily from Collective's delivery of billions of ads on behalf of our clients
every month. Such features may include web site visitation history, geographic
information derived from the IP address, device and browser information
inferred from the user-agent string within the browser HTTP request, and other
data available in the process of ad delivery. This data is both high-dimensional
and sparse. One can think of this data set as an $n \times p$ matrix where each
of the $n$ rows represents a user, and each of the $p$ columns represent a
feature of the users. In the following, we'll interchangeably refer to this
matrix as ``model matrix'' or ``feature matrix''. Since the user features are
independent of the campaigns or the campaign goals, we construct a single global
model matrix for all models, providing the flexibility, however, for any
particular model to add or remove features from the global model matrix. We
address the challenges and our implementation of maintenance and assembly of
the model matrix in detail in Section 3.

%-------------------------------------------------------------------------------

\subsection{Choice of Modeling Methodology}

Collective's audience modeling platform is primarily concerned with predicting
the probability of an event occurring for a user in the future given a vector of
data known about that user at the time of the prediction. The system
described in this paper uses elastic nets to make these predictions, which are
regularized generalized linear models. To train these models, we use the R
\texttt{glmnet} package \cite{glmnetR}, which is an implementation of
the coordinate descent algorithm described in more detail in \cite{glmnet}. In
this section, we will describe elastic nets at a high level, and then explore
the strengths and limitations of these algorithms for our application.

Let our response variable be denoted by $Y = \{0, 1\}$, where $1$ indicates that
the event occurred in the future, and $0$ indicates that the event did not
occur. Further, let $X$ be an $n \times p$ matrix of data where each row
corresponds to the predictors for a user. Then the elastic net solves for a set
of coefficients for the logistic regression model of the form
\begin{displaymath}
\log \left( \frac {Pr(Y = 1 | x)} {Pr(Y = 0 | x)} \right) = \beta_0 + x^T
\beta.
\end{displaymath}
That is, the log of the odds of an event occurring is expressed as a linear
combination of the features in $X$ and a coefficient vector $\beta$.
Traditionally the coefficients for a generalized linear model are determined by
identifying the vector $\beta$ that maximizes the log likelihood equation
\begin{displaymath}
l(\beta_0, \beta) = \frac{1}{N}\sum_{i = 1}^N y_i \cdot 
(\beta_0 + x_i^T \beta) - \log (1 + e^{\beta_0 + x_i^T \beta}) .
\end{displaymath}
However, when $p$ is large (many predictors) in proportion to $n$ (number of
observations), maximizing the likelihood directly risks significant overfitting
and poor out of sample model performance. Thus, the elastic net seeks
coefficients that minimize
\begin{displaymath}
\min_{(\beta_0, \beta) \in \mathbb R^{p+1}} 
l(\beta_0, \beta) + \lambda P_\alpha(\beta) ,
\end{displaymath}
where $\lambda$ is a regularization parameter, and $P_\alpha(\beta)$ is a
coefficient penalty term defined as
\begin{displaymath}
P_\alpha(\beta) = (1 - \alpha) \frac{1}{2}\|\beta\|^2_{\ell_2} + 
\alpha \| \beta \|_{\ell_1}
\end{displaymath}
for a parameter $\alpha$ satisfying $0 \le \alpha \le 1$. The choice of 
$\alpha = 0$ corresponds to using a ridge regression penalty function, and the
choice of $\alpha = 1$ corresponds to using a lasso penalty function. Any 
intermediate value of $\alpha$ is a compromise between the two. In practice, we
found a value of $\alpha = 0.1$ to provide optimal out-of-sample performance for
our models with limited variation across datasets or time.

We chose the R \texttt{glmnet} implementation of coordinate gradient descent for
elastic nets for a variety of reasons. First, elastic nets perform well on out
of sample data, even when compared to much more sophisticated methods like
random forests and gradient boosting machines. In part, this is due to the high
dimensionality, sparseness and noisiness of the data that we are using. Second,
elastic nets produce models which are easy to predict quickly at massive scale,
due to their relative simplicity. Third, the \texttt{glmnet} implementation is
both easy to use, and is incredibly fast and memory efficient. Table 1 lays out
the specific strengths and limitations of this choice, and we explain the most
significant strengths and limitations below.

\textbf{Logistic regression model (Strength): Produces probabilities.}
The logistic regression model produces probabilities for success outcomes
directly. These probabilities can be helpful in interpreting the model
predictions directly, but more importantly can be used in downstream
applications. For example, groups of users can be created that exceed
probability thresholds, and the probabilities can be used directly in ad serving
or real time bidding applications.

\textbf{Logistic regression model (Strength): Speed of predictions.}
Every day our system must score every user for every model still in production.
Given that we have $\sim$200 million users, and $\sim$1,000 models in production
at a given time, that means we are making over 200 billion predictions every
day. The required computation for a given user and model is a sum-product of the
coefficients and the user features, which can be implemented very easily and
performed incredibly quickly. These 200 billion predictions are done in under an
hour in our current implementation.

\textbf{Logistic regression model (Limitation): Nonlinearity must be specified.}
Because logistic regression assumes linearity in the additive terms of the
log-odds, any more complex nonlinear relationships between predictors and the
log-odds must be parameterized in the model matrix in advance. We do so by
either binning continuous features, or by using nonlinear transformations such
as polynomials or splines. In general, the majority of our predictor data is
categorical, and so this is not a material concern for our application.

\textbf{Logistic regression model (Limitation): Interactions must be specified.}
Because logistic regression assumes additivity of terms in the log-odds, any
interactions effects must be specified in the model matrix in advance. With a
large number of features, this quickly becomes intractable in both processing
time and memory storage, so interactions must be added sparingly.

For this application, we believe that interaction terms are less important than
they might be for other applications. First, we are only capturing data related
to users and their past behavior. This means all of our data is directly related
to a single entity (the user), and so complex interactions between multiple
entities are not present. Second, many of our features are sparsely populated,
and so interactions of those sparse features are themselves highly sparse, and
necessarily less likely to influence the prediction. Finally, we have a very
large number of features, drawn from many disparate sets of data. We believe
that useful interaction effects amongst features become less useful as
additional features are added that can explain those interactions directly.

\textbf{Elastic net regularization (Strength): Reduced risk of overfitting.}
Regularization is a crucial feature of elastic nets. By penalizing coefficient
sizes, we ensure that the models are much more likely to generalize to new data,
and thus perform well in production. This enables us to more freely add large
sets of sparse features without being concerned that small sample size models
will overfit them. We also reduce the need to use variable selection techniques
as a part of our daily model building processes, which can be computationally
intensive.

\textbf{Coordinate gradient descent (Strength): $\lambda$ search path for
``free''} In a production machine learning system, one key consideration is the
need to perform grid searches for algorithm meta parameters, which can be
computationally costly. The choice of $\lambda$ in elastic nets has a tremendous
impact on the resulting model, with $\lambda = 0$ producing the (overfitting)
full GLM fit, and $\lambda = \infty$ producing the (underfitting) constant
model. With a wide variety of dataset sizes and outcomes to predict, we find
that in practice optimal (maximum cross-validated AUC) $\lambda$ values
regularly vary from 0.0001 to 0.1. One significant advantage of the coordinate
gradient descent algorithm is that it iteratively solves for the optimal
coefficients for $\lambda$ along a $\lambda$ search path (from largest to
smallest), producing the grid search for this parameter automatically, often faster than if
the $\lambda$ search was not done at all.

\textbf{Coordinate gradient descent (Strength): Sparse data support.} A further
advantage of the coordinate gradient descent algorithm is its support for
sparse matrices. On average, entries in our user matrices are non-zero less than
10\% of the time, and so sparse matrix support reduces the required amount of
memory by a factor of 10. This allows us to use larger matrices that produce
models with higher predictive performance.

\textbf{R package (Strength): Speed of model building.} The R \texttt{glmnet}
package is a wrapper for a Fortran function which implements the coordinate
gradient descent algorithm. This code is highly optimized for speed and is
generally stable in our production system. We build tens of thousands of models
every week, and so this is a significant computational cost improvement over
traditional regularized glm implementations.

\textbf{R package (Limitation): Not parallelized.} The most significant weakness
of this package is that the code is not parallelized. A small amount of
parallelization can be achieved by conducting the cross-validation model builds
in parallel, but in practice we find that we are building many such models, so
parallelization is assumed at the system level across thousands of models run
simultaneously.

\section{Feature Matrix}

Preparing the data for use as predictors in modeling has its own challenges.
First, when the data is captured, it is generally in the format of the source
system and needs to be transformed into data structures that are compatible with
the modeling algorithms. Second, dimensionality of the data needs to be reduced
to address both the extreme sparsity of the data and the scaling of the modeling
algorithms with the number of dimensions. Third, the time structure of the data
requires that the features defined at one point in time could be used for
constructing the data set at another point in time. Finally, the features should
be defined, added, and removed without having to modify the source code. This
section describes Collective's implementation of the data assembly process that
constructs the features used as predictor variables in modeling, as well as
approaches we take to selecting the feature sets.
% As noted above, we will use the terms ``feature matrix'' and ``model matrix''
% interchangeably.

%-------------------------------------------------------------------------------

\subsection{Feature Types}

For our application, the algorithm of choice for building predictive models is
\texttt{glmnet}, as outlined in the previous section. This choice imposes
certain requirements on the structure of the predictor variables. We implemented
four generic types of transformations that convert the features from the format
in which they are available for modeling to final predictor variables that
depend on the specific data at the time when these variables are defined.

Continuous features may need to be transformed to binary format using binning to
account for nonlinearity. In general, binning attempts to break a set of ordered
values into evenly distributed groups, such that each group contains
approximately the same number of values from the sample. In practice, one has to
account for common cases when a disproportionate number of values are the same,
or when the distribution of values is discrete and heavily skewed. Standard
implementations of binning, such as computing quantiles, don't perform
consistently in such cases. An additional consideration is that in the context
of modeling it is impractical to create a bin with very few data points because
that leads to a feature that is extremely sparse and is unlikely to be valuable.
We implemented a robust method for binning that performs well across a wide
variety of distributions and edge cases. This method is minimizing the least
mean square deviation of the resulting number of points in each bin from an
ideal split, where each bin has the same number of points, using a combination
of quantiles and the operations of splitting and merging the bins. The
optimization is constrained by the desired number of bins and by the minimum
number of points in a single bin.

Categorical features must be transformed to binary format by creating a binary
variable for each categorical value. However, high-cardinality features, such as
web pages where ads were delivered, would then translate into millions of
predictor variables, the vast majority of which are extremely sparse. To
alleviate this issue, we apply two approaches. The first is to group related
categorical values. For example, web pages could be grouped by Internet domain,
or by the category of the page contents. The second is to limit the categorical
values that become predictor variables in the feature matrix to only the most
common ones in some sense. 

In our implementation, a categorical variable can be limited by one of two
methods: top coverage and minimum support. The first method, top coverage, is
selecting categorical values by computing the count of distinct users for each
value, sorting the values in descending order by the count of users, and
choosing the top values from the resulting list such that the sum of the
distinct user counts over these values covers $c$ percent of all users, for
example, selecting top geographic locations covering 99\% of users. This works
best with features that only allow one value per user. The minimum support
method is selecting categorical values such that at least $c$ percent of users
have this value, for example, web sites that account for at least $c$ percent of
traffic. This restriction is most amenable to features with more than one
categorical value per user.

To summarize, the first three transformations of the predictor variables are top
coverage and minimum support for categorical variables, and binning for
continuous variables. The fourth and final transformation is a trivial one of
identity, where the predictor variables are taken without change, for example,
when a feature is binary to begin with.

%-------------------------------------------------------------------------------

\subsection{Dimensionality Reduction}

In a typical month, Collective's ad delivery systems encounter billions of
unique browser cookies across millions of online content items. Producing a
feature matrix for every user (cookie) and every piece of information about a
user as an $n \times p$ matrix, where $n$ is the number of users and $p$ is the
number of predictor variables, is impractical both from the data processing
standpoint and because the resulting matrix would only have about 1 in 100,000
non-zero elements. In our modeling system we reduce dimensionality in both $n$
and $p$ to arrive at a few hundred million relevant users and between one and
two thousand predictor variables, with data sparsity of about 1 in 10. The final
feature matrix $X$ is stored in the sparse representation in a database table,
where each row contains the user id $i$, the feature matrix column id $j$, and
the value $X_{ij}$. This dramatically reduces the data storage and processing
requirements, while representing the feature matrix in the format native to
\texttt{glmnet}.

Reduction of the number of predictor variables is achieved primarily by first
selecting the features that become part of the data set, and then limiting the
number of resulting columns in the feature matrix with the transformations
described above. Reduction of the number of users for which we compute the
features arises from the fact that for a large proportion of users we only
deliver a single ad impression (they are either blocking or deleting cookies).
We therefore limit the user universe to those relevant users that have a
reasonable chance of being encountered in the next time period. The relevant
user definition is evolving and is outside the scope of this paper. It is a
balance between potential reach (how many users could be targeted), actual reach
(how many users of those that are targeted will be seen on any given day), and
the requirements for storage and processing power. For example, one could
consider users with ad impressions at least $\Delta t$ apart and vary the time
interval $\Delta t$ to arrive at an acceptable user universe.

%-------------------------------------------------------------------------------

\subsection{Time-Dependent Feature Definitions}

While the data sources that provide the features used for building predictive
models change infrequently, the specific definitions of predictor variables in
the feature matrix depend on the actual data available at the time when the
models are built, because of the need for binning the values of continuous
variables and for limiting the number of values for categorical variables as
described above. If we chose to continuously update the feature matrix
definitions, we would have to rebuild every model in our system in order to make
predictions using the most current definitions. However, we found that once a
sufficient training data sample is available, retraining the models doesn't add
much to the model accuracy, while consuming a lot of resources. 

In our modeling platform, we chose to use the feature definitions computed at
the end of the most recent user observation period, i.e., we update them every
$r$ days, where $r$ is the length of the response interval, typically, once a
week. We rebuild all models at that time as well, and make predictions using
these models every day. The feature definitions comprise a set of source
features included in the feature matrix, the type of each feature, and the
metadata describing the feature transformation into the final predictor
variables. The feature transformation metadata includes the bin boundaries and
the categorical value sets, along with the assignment of each of the predictor
variables to a specific column of the feature matrix. The definitions computed
at a given time are used for generating the feature matrix for all users until
the next update, as well as for generating the feature matrix for each of the
past intervals to assemble the training sets for the modeling. This reprocessing
of past intervals' data is computationally demanding but necessary to ensure all
training data is consistent.

%-------------------------------------------------------------------------------

\subsection{Feature Selection}

The choice of \texttt{glmnet} as a modeling algorithm means that feature
selection is not of immediate importance in ensuring a given model generalizes
well, as regularization ensures that coefficients for features unrelated to the
response are close to zero. However, we still have to choose from amongst
hundreds of thousands of potential features a subset to include in the modeling
platform. This choice requires insight into which of the features have the
greatest impact upon model performance. Further, such insight can be invaluable
in steering feature engineering decisions and data collection decisions in the
broader technology platform over time.

Absent regularization, ANOVA tests can be used to compute p-values for terms
included in a generalized linear model. Assuming the system that generated the
data conforms to the assumptions of the model (rare in our experiences), these
p-values can be a reliable way of identifying which features are most
significant. For lasso models, it is possible to compute the covariance test
statistic \cite{tibshirani:lasso}. However, this is a recent development that
has not been generalized to elastic nets.

Instead, we employ two different approaches for estimating the importance of
features. The first is commonly referred to as a dropterm, wherein we group sets
of related features (e.g., all geographic features) and compare the predictive
performance of a model without those features (the 'dropped' model) to one with
them (the 'full' model). We perform 5-fold cross validation to both the full and
dropped models and compare their area under the curve (AUC) statistics. Feature
groups whose removal does not materially reduce the AUC are considered to be
'weak' feature groups for a given model. We evaluate both the average change in
AUC across all models in production as well as the distribution in change in AUC
as some features will be highly important for a small number of models, but not
important for the majority of others.

The above approach is too computationally intensive to run continuously in
production. It requires building every model five times (for cross-validation)
for every feature group, which can be in the 100s depending on the level of
grouping performed. So in practice this is performed periodically on a
representative sample of models as significant changes to the set of features is
contemplated by our data sciences team.

However, we still wish to have a measure of variable importance continuously
available in our production system. This is useful both to report on insights
for individual models, and also to track any changes over time in feature group
performance which might indicate upstream data availability or processing
issues. We have found that a simplified approach to measuring the impact of a
feature group on predictions to be correlated to the more robust dropterm
approach described above but far less computationally intensive. Specifically,
for every group of features we set their coefficients to zero in each model, and
compare the assignments made by the altered model to those made by the original
model. Given that the models are regularized, this tests whether or not the
coefficients are meaningfully altering the predictions, without being overly
sensitive to highly correlated data (which could be the case in an unregularized
GLM).

%-------------------------------------------------------------------------------

\subsection{Assembly}

The main requirements for the system that assembles the feature matrix were to
(i) define features for a given set of models without modifying the source code
of the system; (ii) maintain multiple sets of features for different sets of
models; (iii) persist and maintain multiple feature definitions depending on the
time when they were generated; and (iv) assemble the feature matrix on demand
for a given subset of users, the feature definitions, and the source data as of
a specific date. We separated the feature matrix generation into three stages:
setting up configuration, computing feature definitions, and producing an
instance of the feature matrix. The logical separation of these stages provides
the required flexibility to satisfy the requirements in production, and allows
for further research and exploration of the features as new data sets become
available.

First, we define which data becomes part of the feature matrix. This is done
through manually setting up a configuration that defines the source of every
feature, the feature transformation type, parameters of the transformation, as
well as other metadata that is useful in reporting and visualization. All data
sources are pre-processed and assembled in a relational database (see the
section on Systems Architecture below for more detail). The feature data source
is most commonly a database table containing the user ids and a column with the
feature values. The configuration allows for specifying custom transformations
of the column values using SQL, as well as for applying filters that limit the
rows included in the data set to those that satisfy the filter conditions. The
parameters of the feature transformations described above include, for example,
the desired number of bins and the minimum percent of values in each bin for the
binning algorithm, and the percent $c$ parameter for the top coverage and the
minimum support transformations. The configurations are versioned, and we
maintain separate ones for modeling user behavior in different countries.

The second stage is the generation of the specific feature matrix definitions at
a particular time. These definitions rely on the data collected during the most
recent user observation period and available at the end of that period. These
definitions are also versioned for each configuration. We maintain a history of
the feature matrix definitions so that we are able to track the changes of the
definitions in time and perform system diagnostics and troubleshooting. As
discussed above, the feature matrix definitions are updated every $r$ days,
where $r$ is the length of the response interval.

The third and final stage is the generation of an instance of the feature matrix
using the given feature matrix definitions and the date as of which the features
data was available. We generate an instance of the feature matrix using the most
recent definitions and use it for making predictions every day. In addition,
during the process of assembly of the training data sets for building models, we
generate the feature matrix on demand just for users that are part of the data
sample used for modeling. The latter usually happens to be a small fraction of
all users because of the sparsity of the positive responses and downsampling of
the negative responses.

% Figure 3
\begin{figure}
\centering
\scalebox{0.8}{\input{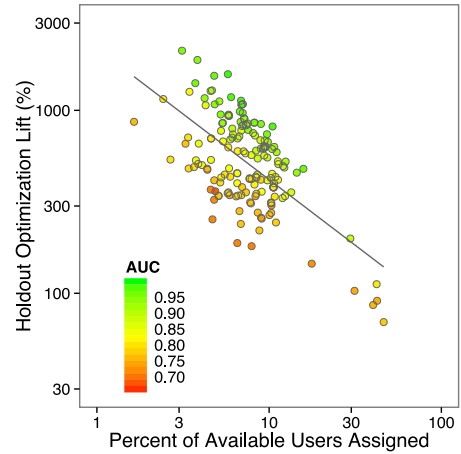}}
%\scalebox{0.5}{\includegraphics{holdout_lift/holdout_lift}}
%\scalebox{0.5}{\includegraphics{try}}
\caption{Evaluating model performance.}
\end{figure}

% Figure 4
\begin{figure*}
\centering
\input{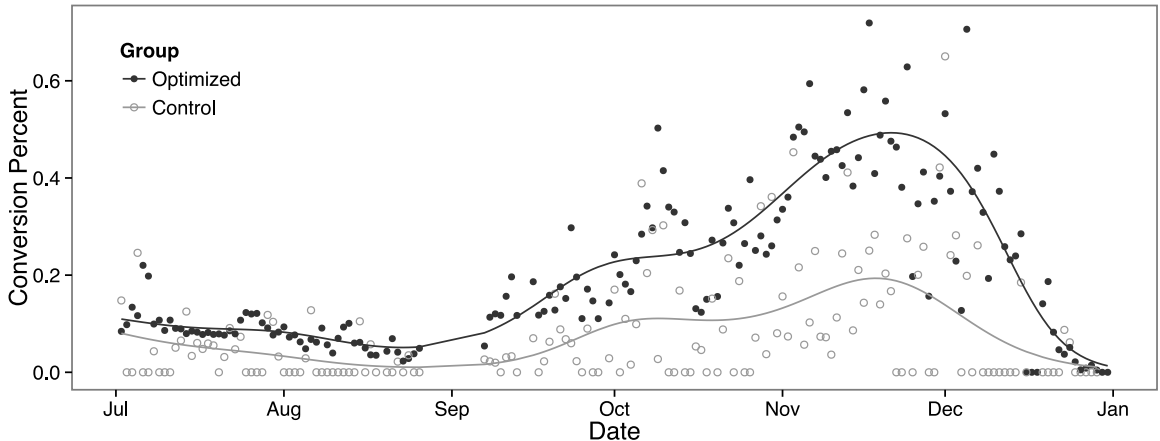}
\caption{Sample model report of performance of optimized and control groups in
production.}
\end{figure*}

\section{Model Evaluation}

A specific application of Collective's audience modeling platform described in
this paper is selecting a target audience for each campaign and performance
goal. This is achieved by making predictions for every user in a large universe
of users, sorting the predicted probabilities in descending order and assigning
the $N$ highest scoring users to the target audience. The size of the target
audience $N$ is a balance between accuracy of the predictions and the possible
reach of the campaign.  

To align the model evaluation methodology with the practical application at
hand, we introduce the concept of ``optimization lift''. We define the
optimization lift $\mathcal{L}$ as
\begin{displaymath}
\mathcal{L} = \frac{ \Pr_N(\mathrm{success} | \mathrm{optimized}) }
{ \Pr_N(\mathrm{success} | \mathrm{random}) } - 1,
\end{displaymath}
where the numerator is the probability of success (true positive rate) in the
optimized set of top $N$ users, and the denominator is the probability of
success in a random set of $N$ users.

Figure 3 illustrates the relationship between $\mathcal{L}$ evaluated on a
contemporaneous holdout set, $N$ and the model AUC. Each point represents a
model, and $\mathcal{L}$ is on the vertical axis and the horizontal axis is
the percent of available users assigned ($N$ over the size of the
relevant universe of users for each model). The color of each point is mapped to
the AUC for the model. This illustration shows that the size of the target
audience assigned has a dramatic effect on the performance of the model.
However, given an assigned percentage of users, there remains close to an order
of magnitude variation in $\mathcal{L}$ driven by the AUC.

In our modeling platform, we take a three-level approach to evaluating the
models. The first level is cross-validation to tune the model meta-parameters.
In case of \texttt{glmnet} we are searching for the optimal value of $\lambda$.
This approach has the advantage of being available at the model build time, but
carries the risk of overfitting the model to the training set. The second
approach is testing the model on a contemporaneous holdout set of data. The
advantage of such testing is that it is available pre-deployment, but the model
may still not generalize in the noisy production environment. We compute and
report the AUC and the optimization lift $\mathcal{L}$ for each model and set up
alerts that send notifications for models that do not achieve a minimally
required accuracy. 

Finally, the third approach is embedding a random control group of users into
each optimized target audience. This allows us to measure actual lift in the
production environment, but it takes time and a sufficient number of impressions
delivered for the campaign to achieve accuracy and statistical significance.
Figure 4 shows a sample performance report, where the black dots and line
represent daily performance of the optimized user group and the grey circles and
line represent the daily performance of the random control group. The average
lift of the optimized target audience relative to the control, computed from a
sample of 58 campaigns from advertisers across different industries over a
period of time in the first half of 2013 was 400\%, i.e., the optimized audience
performed 5 times better than random users.

\section{System Implementation}

\subsection{Core Technologies}

The architecture for a predictive modeling platform carries a number of
constraints. The choice of a modeling algorithm determines which implementations
are available, and that limits further choices. In our case, building the models
themselves had to be done in R once we chose \texttt{glmnet}. Constructing the
data sets for modeling is very intensive in data processing and requires
sufficient capacity for the system. Virtually all of our data is structured and
many relevant data transformations involve joins across data sets. These and
other considerations led us to build the data-centric portion of our modeling
platform on an IBM PureData (formerly, Netezza TwinFin) appliance. At its core is
a parallel relational database with SQL interface, capable of storing and
processing tables with hundreds of billions of rows. Since we needed to use both
R and SQL within the system, we chose to standardize around these two languages.

\subsection{Interaction With End Users}

While the audience modeling process is implemented and maintained by the data
sciences and engineering teams, the main end users of the system are the ad
operations and campaign optimization teams. These teams set up the specific
inputs for each of the models, such as the campaign metadata (e.g., campaign
name), the types of the performance goals (e.g., clicks or conversions), the
conversion pixel identifiers (if applicable), and the desired size of the target
audience. In a separate user interface, the data sciences team sets up the
global parameters of the models as well as the configurations of the feature
matrix. 

Upon completion of every model build, all relevant teams receive several reports
detailing the results. These reports include the AUC and optimization lift for
each model based on the holdout set, and flag any issues that may have occurred.
For example, when a campaign has just started, there may not be enough positive
events to build a meaningful model. This situation is normal, however, the
optimization team receives a corresponding alert in the user interface.

As soon as the campaign has delivered enough impressions so that the
optimization lift based on the embedded control group can be computed with
sufficient statistical significance, a performance report similar to that in
Figure 4 is produced automatically and is made available to the optimization and
account services teams.

\subsection{Modeling Process}

% Figure 5
\begin{figure}
\centering
\scalebox{0.9}{\input{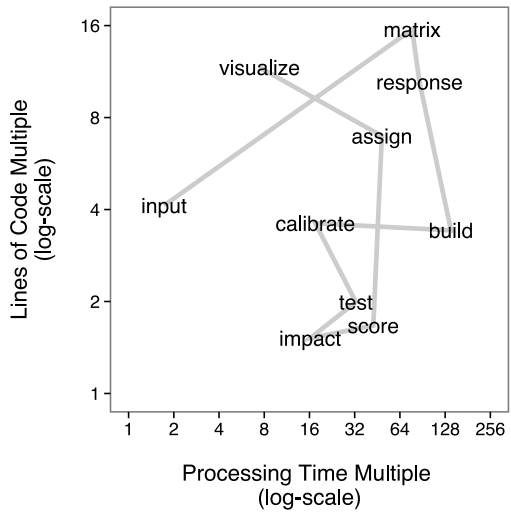}}
\caption{Lines of code involved in each step of the processing.}
\end{figure}

The audience modeling process consists of 10 distinct phases, which are
visualized in Figure 5 in terms of their relative processing time (indicating
computational intensity) and lines of code (indicating implementation
complexity). Note that the most computationally intensive tasks are highly
parallelized, and so in reality are much more computationally intensive than
they appear in the above chart. Each phase is briefly described below.

\textbf{Input.} The input phase is processing and structuring user input, both
specific to individual models, and the meta-data controlling the entire process
($\approx$50 parameters)

\textbf{Matrix.} This phase assembles the feature matrix for training and
prediction, and was described in detail in Section 3.5. 

\textbf{Response.} The response stage assembles the response data for all models
and performs relevant sampling. Here we achieve significant savings in
processing time by taking advantage of similarities in the structure of the data
across model types and performing many of the data transformations on the whole
data at once. There are two types of data sampling that we apply for efficiency.
First, we limit the maximum number of rows in the training data on the basis of
evaluating the impact of adding more data on the model accuracy relative to the
additional resources for extra storage and computation. Second, in the most
common case when the negative events in the response greatly outnumber the
positive (assuming binary response, as described in Section 2.1) we downsample
the negative events to improve speed and avoid potential numerical instability
when building the model.

\textbf{Build.} The build phase involves combining the feature matrix and the
response into the training data sets, distributing the modeling tasks to a
cluster of computers, training the \texttt{glmnet} models, and then collecting
and structuring the results. Although the \texttt{glmnet} implementation in R is
not parallelized, the large number of models that we need to build at a time
allows us to achieve parallelism by building a number of models simultaneously.
For this purpose, we leverage our Hadoop cluster. Note that this is not a
typical big data problem solved with Hadoop. Instead, we treat Hadoop worker
nodes as parallel processors while taking advantage of its architecture. The
modeling jobs are implemented as a mapper-only map-reduce job, where the mapper
is generated by the audience modeling platform and calls R to build a single
model.

\textbf{Calibrate.} This stage provides an optimization by making adjustments to
the model coefficients to speed up the downstream scoring of all users for every
model. Since the final result of the process for each model is a set of $N$
highest scoring users and the model is linear, we adjust the coefficients so
that the top $N$ users have positive scores, while the rest have negative
scores. This adjustment can be estimated, for example, by scoring a sample of
users. As a result, there is no need to sort the complete set of scores for
every model to determine the top $N$ users. Moreover, during scoring only
information about users with positive scores needs to be stored, which leads to
significant savings in the amount of data written to disk.

\textbf{Test.} The test phase performs out-of-sample testing on a contemporaneous
sample of data. 

\textbf{Impact.} The impact phase calculates the impact of each feature group on
predictions by setting the model coefficients for the features in the group to
zero and comparing the resulting predictions to those made by the unmodified
model. As described in Section 3.4, this serves as a proxy for the full dropterm
feature selection approach.

\textbf{Score.} The scoring phase is making predictions using each model for
each relevant user. Since the \texttt{glmnet} models are linear, a user's score
from a given model is the sum-product of the model coefficients and the values
of the feature matrix for this user. The computation of the scores for multiple
models and all users can be viewed as a matrix multiplication problem: if the
number of models is $q$, the resulting matrix of user scores $S$, of size $n
\times q$ is equal to $X \times C$, where $X$ is the $n \times p$ feature
matrix, and $C$ is the $p \times q$ matrix of model coefficients. Each column of
$C$ contains the model coefficients for a single model. While $n$ is on the
order of hundreds of millions users, both $p$ and $q$ are on the order of 1000.
Thus, the scoring problem is equivalent to multiplying a matrix on the order of
100 million rows by 1000 columns with another matrix of size about $1000 \times
1000$.

Although matrix multiplication is well understood, the naive approaches did not
perform well in this case. In our implementation of scoring, we take
advantage of several factors. First, because the matrix of coefficients $C$ is
relatively small, one can replicate it on multiple nodes of a parallel
architecture, and compute scores for distinct sets of users on each node.
Second, an algorithm that takes into account the fact that both $X$ and $C$ are
sparse can reduce the number of operations required to compute the scores by
orders of magnitude. Finally, as we noted above in the description of the
calibration phase, one can drastically reduce the time needed to write the
results of calculations to disk by limiting the output to only relevant values.

\textbf{Assign.} The assignment phase includes performing the final assignment
of users to target audiences and delivery of the results to real-time systems
for targeting. At this stage we embed the control group in the target audience.
We also optimize the size of the data assembled for delivery, for example, by
computing and only sending the differences in the user assignment since the last
model build.

\textbf{Visualize.} Each modeling process run concludes with producing a
comprehensive set of visualizations for each phase of the current run, as well
as time series views of prior runs. All visualizations are automated and utilize
the \texttt{ggplot2} R package \cite{ggplot}.

Not surprisingly, the tasks that require the most computational time are either
incredibly data intensive (matrix and response assembly) or CPU intensive (model
building). That the scoring phase is comparatively less computationally
intensive is a testament to how fast glmnet models can be scored, and to
numerous optimizations that we implemented.

The data assembly tasks (matrix and response) are also very logically intensive,
as there are many decisions that must be made in those phases. However, their
position is somewhat inflated by the fact that much of the code is written in
SQL, a relatively verbose language (compared to plyr in R for example). The
visualization phase is in many ways one of the most complex, as we generate
thousands of visualizations for every run, of at least 50 different varieties.

\subsection{Measurement and Monitoring}

For any large scale system to be robust, it needs to implement automated
testing, monitoring of correctness and performance of individual components, as
well as measurement and recording of key metrics over time. It should also fail
gracefully when errors and changes in data inevitably occur. An excellent
overview of the design principles of large predictive modeling systems has been
presented in \cite{m6d:robust}. Below we describe some of the measures
implemented in our modeling platform to ensure robustness and provide for
ongoing improvements.

\textbf{Integration testing.}
When changes are made to such a large and complex system consisting of dozens of
interdependent steps, it is difficult to anticipate all effects on different
parts of the system. We have addressed this by testing the code changes on a
test data set with a few representative models, and verify that the data
produced and the performance of the models on the holdout set are consistent
with the introduced changes.

\textbf{Timing.}
For the whole system to be scalable, each component must run in within
acceptable time limits. We take the development approach where we design each
component to be fast but without excessive complexity, measure their execution
time, and iterate on the bottlenecks. To be able to follow this approach, we
measure and record the timing of every step of the process, down to individual
SQL queries. Because for many tasks we are using R to run SQL queries in the
database, we have implemented wrappers that can automatically record the timing
of every query.

\textbf{Model performance testing}
The models are tested at three levels as described in detail in Section 4.

\textbf{Monitoring key metrics.}
We store a history of the most important measurements and thus make it possible
to monitor the system performance and scaling over time. When key metrics change
unexpectedly, we investigate and take action, as necessary.

\textbf{Error checking.}
We have two key layers of error checking in the system. The first layer is
system-wide, where at every phase of the multi-step process we run multiple
diagnostics against the intermediate data available at that step. These tests
may be as simple relational integrity checks or more complex tests based on
custom logic. Whenever these system-wide checks fail it is indicative of a
systemic issue that typically needs to be addressed in the code. The second
layer is model specific, and captures any errors that arise due to input
inconsistencies or missing or incomplete data associated with a specific model.

%Extensive error checking for data consistency is performed at every
%stage of the modeling process. The system verifies that the assumptions made
%about the input data hold before data transformations are executed, as well as
%that the output data conforms to the expected content and format. This
%includes, for example, checking for nulls, empty tables, expected number of
%rows, etc.

\textbf{Error handling.}
There are multiple severity levels of errors that may occur. First, and the most
severe one, is an error that causes the whole modeling process to halt. This
could be an infrastructure failure, such as a network interruption or disk
failure. These errors are very rare but the most disruptive, as multiple teams
have to get involved to resolve the issue and restart the automated processes.
Second, there are errors resulting from the required input data not being
present, for example, when there is a problem with data transfer from one of the
many data sources required for the data assembly. The system generally deals
with this by waiting for some time for the data to arrive, as well as by sending
out alerts. Finally, there are errors that occur during the modeling process.
These are anticipated, diagnosed and flagged in the system and presented to end
users without disrupting the flow of the overall modeling process. In addition,
in cases of model specific errors, the system may fall back to another relevant
model type to assign the target audience. 

\textbf{Visualization}
In addition to the error checking described above, we automatically generate
hundreds of visualizations covering every stage of the modeling system. These
visualizations help build our intuition for what a ``healthy'' state of the
system looks like, and are thoroughly reviewed anytime a meaningful change is
made to the system to look for introduced anomalies. Having these visualizations
available at the individual model level, at the system component level, as well
as across the whole system, proved to be critical to understanding and
monitoring of the system behavior, proposing improvements, and troubleshooting.

\section{Conclusions}

Building a single accurate and scalable machine learning model to predict
audience behavior for an advertiser given hundreds of terabytes of data covering
hundreds of millions of users and millions of potential features is a challenge.
Architecting, implementing and supporting a system to build thousands of such
models, and making certain that they run daily to ensure the proper delivery of
billions of advertising impressions monthly is even more challenging. Through
careful choices in data assembly, algorithm implementation and system controls
and monitoring the platform described in this paper has enabled the accurate
delivery of billions of advertisements in multiple countries on behalf of
Collective's clients.

In this paper, we focused on predicting user actions in the context of digital
advertising campaigns with performance goals. However, the same modeling
platform, with minor modifications, is working for campaigns with reach goals,
predicting whether a given user is likely to belong to a certain group of
people. In addition, because our models produce probabilities for success
outcomes, the platform has been extended to generate inputs for calculation of
bid amounts for buying ad impressions at real-time bidding (RTB) ad exchanges.
The described system can be further generalized to applications beyond digital
advertising, in any situation where one aims to predict user behavior with
multiple, possibly interdependent, outcomes. Such applications may include, for
example, e-commerce and digital publishing.

\section{Acknowledgments}

We thank Ravi Mody and Chris Bethel of the Collective Data Sciences team for
their helpful feedback on this manuscript.

%
% The following two commands are all you need in the
% initial runs of your .tex file to
% produce the bibliography for the citations in your paper.
\bibliographystyle{abbrv}
\bibliography{modeling_at_scale}  

\begin{thebibliography}{10}

\bibitem{criteo:scalable}
O.~Chapelle, E.~Manavoglu, and R.~Rosales.
\newblock A simple and scalable response prediction for display advertising.
\newblock {\em ACM Trans. Intell. Syst. Technol.}, 2014.
\newblock To appear.

\bibitem{turn:convrate}
K.~chih Lee, B.~Orten, A.~Dasdan, and W.~Li.
\newblock Estimating conversion rate in display advertising from past
  performance data.
\newblock In {\em KDD '12 Proceedings of the 18th ACM SIGKDD international
  conference on Knowledge discovery and data mining}. SIGKDD, ACM Press, 2012.

\bibitem{turn:perfgoal}
A.~Dasdan and A.~Svirsky.
\newblock Model-based performance goal adjustment in real-time online
  advertising campaigns.
\newblock In {\em KDD '12 Proceedings of the 18th ACM SIGKDD international
  conference on Knowledge discovery and data mining}, pages 804--812. SIGKDD,
  ACM Press, 2012.

\bibitem{glmnetR}
J.~Friedman, T.~Hastie, and R.~Tibshirani.
\newblock glmnet: Lasso and elastic-net regularized generalized linear models.
\newblock \url{cran.r-project.org/web/packages/glmnet/}.
\newblock Published: 2013-08-04.

\bibitem{glmnet}
J.~Friedman, T.~Hastie, and R.~Tibshirani.
\newblock Regularization paths for generalized linear models via coordinate
  descent.
\newblock {\em J. Stat. Softw.}, 33(1):1--22, 2010.

\bibitem{IAB_3ms}
Interactive{\ }Advertising{\ }Bureau.
\newblock Making measurement make sense (3ms) initiative.
\newblock \url{http://www.iab.net/MMMS}, 2013.

\bibitem{m6d:bidopt}
C.~Perlich, B.~Dalessandro, R.~Hook, O.~Stitelman, T.~Raeder, and F.~Provost.
\newblock Bid optimizing and inventory scoring in targeted online advertising.
\newblock In {\em KDD '12 Proceedings of the 18th ACM SIGKDD international
  conference on Knowledge discovery and data mining}, pages 804--812. SIGKDD,
  ACM Press, 2012.

\bibitem{m6d:robust}
O.~Stitelman, B.~Dalessandro, C.~Perlich, and F.~Provost.
\newblock Design principles of massive, robust prediction systems.
\newblock In {\em KDD '12 Proceedings of the 18th ACM SIGKDD international
  conference on Knowledge discovery and data mining}, pages 1357--1365. SIGKDD,
  ACM Press, 2012.

\bibitem{tibshirani:lasso}
R.~J. Tibshirani, R.~J. Tibshirani, R.~Lockhart, and J.~Taylor.
\newblock A significance test for the lasso.
\newblock {\em Annals of Statistics}, 2014.
\newblock To appear.

\bibitem{ggplot}
H.~Wickham.
\newblock {\em ggplot2: Elegant Graphics for Data Analysis}.
\newblock Springer, New York, 2009.

\end{thebibliography}
% You must have a proper ".bib" file
%  and remember to run:
% latex bibtex latex latex
% to resolve all references
%
% ACM needs 'a single self-contained file'!
%
%APPENDICES are optional
%\balancecolumns
\end{document}